\documentclass[10pt,twocolumn,letterpaper]{article}

\usepackage{cvpr}
\usepackage{times}
\usepackage{epsfig}
\usepackage{graphicx}
\usepackage{amsmath}
\usepackage{amssymb}
\usepackage{booktabs}
\usepackage[dvipsnames]{xcolor}
\usepackage{pbox}
\usepackage{pifont}
\newcommand{\cmark}{\ding{51}}%

{

\usepackage[pagebackref=true,breaklinks=true,letterpaper=true,colorlinks,bookmarks=false]{hyperref}

\cvprfinalcopy 

\def\cvprPaperID{xxx} 

\definecolor{myblue}{RGB}{81,167,249}

\begin{document}


\title{SuperPoint: Self-Supervised Interest Point Detection and Description}

\author{Daniel DeTone\\
Magic Leap\\
Sunnyvale, CA\\
{\tt\small ddetone@magicleap.com}
\and
Tomasz Malisiewicz\\
Magic Leap\\
Sunnyvale, CA\\
{\tt\small tmalisiewicz@magicleap.com}
\and
Andrew Rabinovich\\
Magic Leap\\
Sunnyvale, CA\\
{\tt\small arabinovich@magicleap.com}
}

\maketitle

\begin{abstract}
This paper presents a self-supervised framework for training interest point detectors and descriptors suitable for a large number of multiple-view geometry problems in computer vision. As opposed to patch-based neural networks, our fully-convolutional model operates on full-sized images and jointly computes pixel-level interest point locations and associated descriptors in one forward pass. We introduce Homographic Adaptation, a multi-scale, multi-homography approach for boosting interest point detection repeatability and performing cross-domain adaptation (\eg, synthetic-to-real). Our model, when trained on the MS-COCO generic image dataset using Homographic Adaptation, is able to repeatedly detect a much richer set of interest points than the initial pre-adapted deep model and any other traditional corner detector. The final system gives rise to state-of-the-art homography estimation results on HPatches when compared to LIFT, SIFT and ORB.

\end{abstract}


\vspace{-.2in}
\section{Introduction}
\vspace{-.05in}
The first step in geometric computer vision tasks such as Simultaneous Localization and Mapping (SLAM), Structure-from-Motion (SfM), camera calibration, and image matching is to extract interest points from images. Interest points are 2D locations in an image which are stable and repeatable from different lighting conditions and viewpoints. The  subfield of mathematics and computer vision known as Multiple View Geometry~\cite{hartley2003multiple} consists of theorems and algorithms built on the assumption that interest points can be reliably extracted and matched across images. However, the inputs to most real-world computer vision systems are raw images, not idealized point locations.

\begin{figure}[t]
\begin{center}
\includegraphics[width=1.0\linewidth]{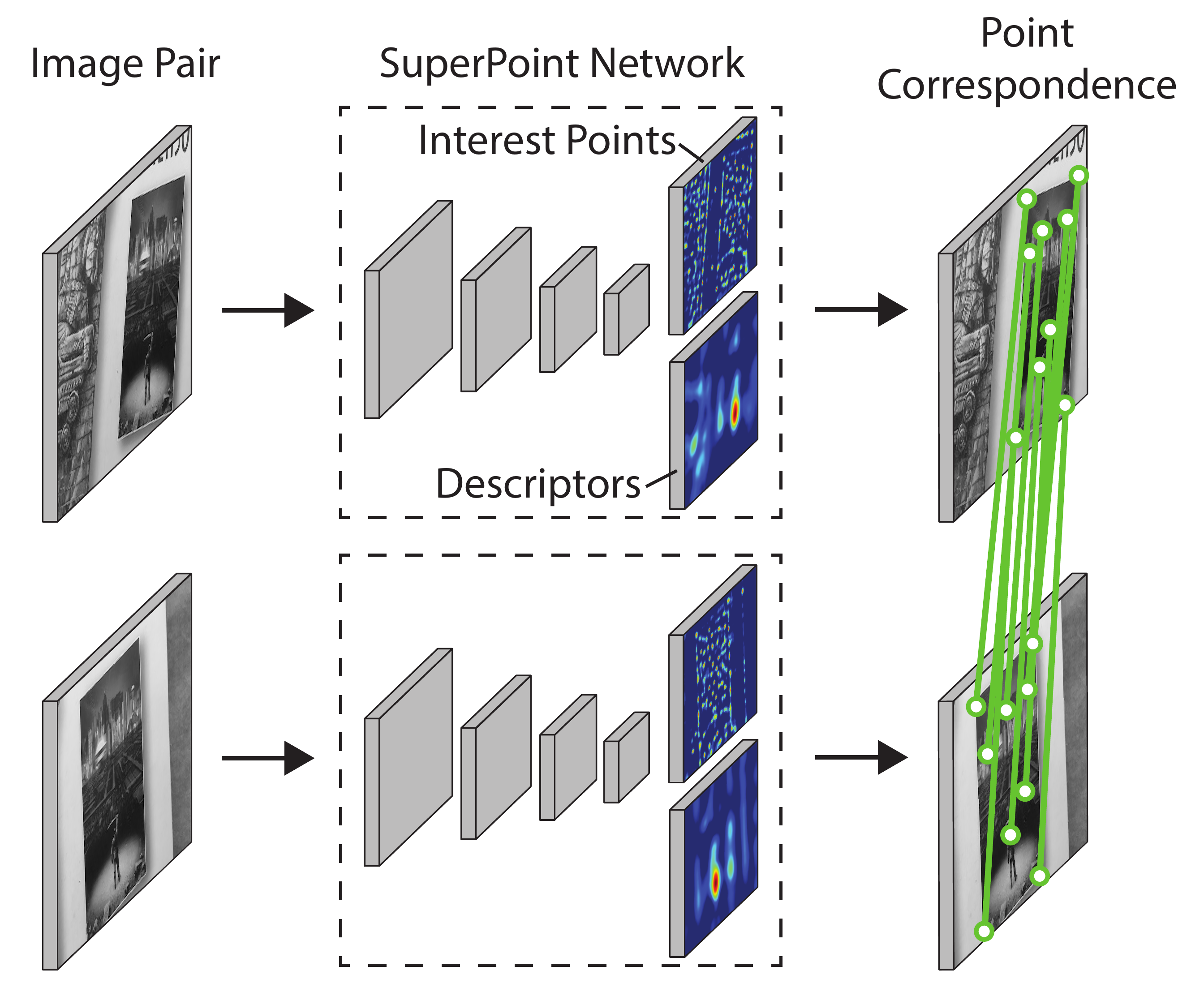}
\end{center}
\vspace{-.1in}
\caption{{\bf SuperPoint for Geometric Correspondences.} We present a fully-convolutional neural network that computes SIFT-like 2D interest point locations and descriptors in a single forward pass and runs at 70 FPS on $480\times640$ images with a Titan X GPU. \label{fig:teaser}} 
\vspace{-.2in}
\end{figure}

Convolutional neural networks have been shown to be superior to hand-engineered representations on almost all tasks requiring images as input. In particular, fully-convolutional neural networks which predict 2D ``keypoints'' or ``landmarks'' are well-studied for a variety of tasks such as human pose estimation~\cite{wei2016cpm}, object detection~\cite{LiuAESR15}, and room layout estimation~\cite{lee2017roomnet}. At the heart of these techniques is a large dataset of 2D ground truth locations labeled by human annotators. 

\begin{figure*}[t]
\begin{center}
\vspace{-.5in}
\includegraphics[width=1.0\linewidth]{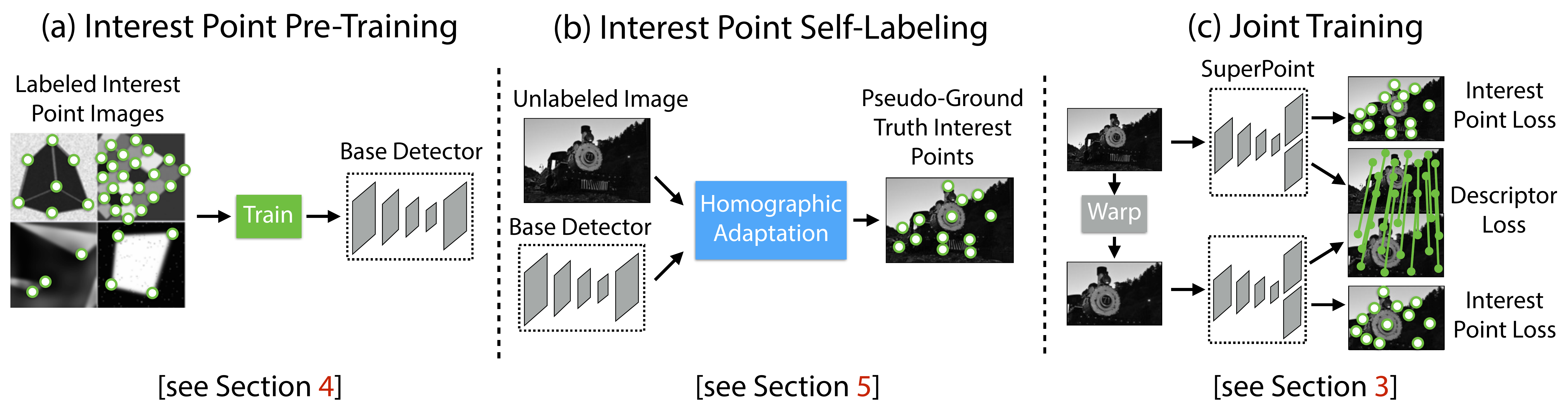}
\vspace{-.3in}
\end{center}
\caption{{\bf Self-Supervised Training Overview.} In our self-supervised approach, we (a) pre-train an initial interest point detector on synthetic data and (b) apply a novel Homographic Adaptation procedure to automatically label images from a target, unlabeled domain. The generated labels are used to (c) train a fully-convolutional network that jointly extracts interest points and descriptors from an image. \label{fig:self_supervised}}
\vspace{-.1in}
\end{figure*}

It seems natural to similarly formulate interest point detection as a large-scale supervised machine learning problem and train the latest convolutional neural network architecture to detect them. Unfortunately, when compared to semantic tasks such as human-body keypoint estimation, where a network is trained to detect body parts such as the corner of the mouth or left ankle, the notion of interest point detection is semantically ill-defined. Thus training convolution neural networks with strong supervision of interest points is non-trivial.

Instead of using human supervision to define interest points in real images, we present a self-supervised solution using self-training. In our approach, we create a large dataset of pseudo-ground truth interest point locations in real images, supervised by the interest point detector itself, rather than a large-scale human annotation effort. 

To generate the pseudo-ground truth interest points, we first train a fully-convolutional neural network on millions of examples from a synthetic dataset we created called \emph{Synthetic Shapes} (see Figure~\ref{fig:self_supervised}{\color{red}a}). The synthetic dataset consists of simple geometric shapes with no ambiguity in the interest point locations. We call the resulting trained detector \emph{MagicPoint}---it significantly outperforms traditional interest point detectors on the synthetic dataset (see Section~\ref{sec-mp}). MagicPoint performs surprising well on real images despite domain adaptation difficulties~\cite{ganin2015unsupervised}. However, when compared to classical interest point detectors on a diverse set of image textures and patterns, MagicPoint misses many potential interest point locations. To bridge this gap in performance on real images, we developed a multi-scale, multi-transform technique $-$ \emph{Homographic Adaptation}.

Homographic Adaptation is designed to enable self-supervised training of interest point detectors. It warps the input image multiple times to help an interest point detector see the scene from many different viewpoints and scales (see Section~\ref{sec-ha}). We use Homographic Adaptation in conjunction with the MagicPoint detector to boost the performance of the detector and generate the pseudo-ground truth interest points (see Figure~\ref{fig:self_supervised}{\color{red}b}). The resulting detections are more repeatable and fire on a larger set of stimuli; thus we named the resulting detector \emph{SuperPoint}.

The most common step after detecting robust and repeatable interest points is to attach a fixed dimensional descriptor vector to each point for higher level semantic tasks, \eg, image matching. Thus we lastly combine SuperPoint with a descriptor subnetwork (see Figure~\ref{fig:self_supervised}{\color{red}c}). Since the SuperPoint architecture consists of a deep stack of convolutional layers which extract multi-scale features, it is straightforward to then combine the interest point network with an additional subnetwork that computes interest point descriptors (see Section~\ref{sec-sp}). The resulting system is shown in Figure~\ref{fig:teaser}.



\vspace{-.05in}
\section{Related Work}
\vspace{-.05in}
\label{sec-related}

Traditional interest point detectors have been thoroughly evaluated~\cite{Schmid2000, Mikolajczyk2005}. The FAST corner detector~\cite{rosten2006} was the first system to cast high-speed corner detection as a machine learning problem, and the Scale-Invariant Feature Transform, or SIFT~\cite{lowe2004}, is still probably the most well-known traditional local feature descriptor in computer vision.

Our SuperPoint architecture is inspired by recent advances in applying deep learning to interest point detection and descriptor learning. At the ability to match image sub-structures, we are similar to UCN~\cite{choy_nips16} and to a lesser extent DeepDesc~\cite{serra15}; however, both do not perform any interest point detection. On the other end, LIFT~\cite{yi16}, a recently introduced convolutional replacement for SIFT stays close to the traditional patch-based detect then describe recipe. The LIFT pipeline contains interest point detection, orientation estimation and descriptor computation, but additionally requires supervision from a classical SfM system. These differences are summarized in Table~\ref{tbl:novelty}.

\begin{table}[h]
  \vspace{-.05in}
  \centering
  \scriptsize
  \setlength{\tabcolsep}{2pt}
  \begin{tabular}{c|c|c|c|c|c}
          & \pbox{4cm}{Interest \\ Points?} & \pbox{4cm}{Descriptors?} & \pbox{4cm}{Full Image \\ Input?} & \pbox{4cm}{Single\\Network?} & \pbox{4cm}{Real\\Time?} \\
          \hline
		  SuperPoint (ours) & \cmark & \cmark & \cmark & \cmark & \cmark \\
          \hline
          LIFT \cite{yi16} & \cmark & \cmark &  &  &  \\
          \hline
          UCN \cite{choy_nips16} & & \cmark & \cmark & \cmark & \\
          \hline
          TILDE \cite{verdie15} & \cmark & & & \cmark & \\
          \hline
          DeepDesc \cite{serra15} & & \cmark & & \cmark & \\
          \hline
          SIFT  & \cmark & \cmark &  & & \\
          \hline
          ORB  & \cmark & \cmark & & & \cmark \\
          \hline
  \end{tabular}
  \vspace{-.1in}
  \bigskip
  \caption{\textbf{Qualitative Comparison to Relevant Methods.} Our SuperPoint method is the only one to compute both interest points and descriptors in a single network in real-time. }
  \label{tbl:novelty}
\vspace{-.15in}
\end{table}

On the other extreme of the supervision spectrum, Quad-Networks~\cite{Savinov16} tackles the interest point detection problem from an unsupervised approach; however, their system is patch-based (inputs are small image patches) and relatively shallow 2-layer network. The TILDE~\cite{verdie15} interest point detection system used a principle similar to Homographic Adaptation; however, their approach does not benefit from the power of large fully-convolutional neural networks.

Our approach can also be compared to other self-supervised methods, synthetic-to-real domain-adaptation methods. A similar approach to Homographic Adaptation is by Honari \etal~\cite{Honari2017} under the name ``equivariant landmark transform.'' Also, Geometric Matching Networks~\cite{Rocco2017} and Deep Image Homography Estimation~\cite{detone16} use a similar self-supervision strategy to create training data for estimating global transformations. However, these methods lack interest points and point correspondences, which are typically required for doing higher level computer vision tasks such as SLAM and SfM. Joint pose and depth estimation models also exist~\cite{zhou2017,Vijayanarasimhan17,ummenhofer2017}, but do not use interest points.



\section{SuperPoint Architecture}
\label{sec-sp}

We designed a fully-convolutional neural network architecture called SuperPoint which operates on a full-sized image and produces interest point detections accompanied by fixed length descriptors in a single forward pass (see Figure~\ref{fig:spnet}).  The model has a single, shared encoder to process and reduce the input image dimensionality. After the encoder, the architecture splits into two decoder ``heads'', which learn task specific weights -- one for interest point detection and the other for interest point description. Most of the network's parameters are shared between the two tasks, which is a departure from traditional systems which first detect interest points, then compute descriptors and lack the ability to share computation and representation across the two tasks.

\subsection{Shared Encoder}

Our SuperPoint architecture uses a VGG-style~\cite{vgg} encoder to reduce the dimensionality of the image. The encoder consists of convolutional layers, spatial downsampling via pooling and non-linear activation functions. Our encoder uses three max-pooling layers, letting us define $H_c = H / 8$ and $W_c = W / 8$ for an image sized $H \times W$. We refer to the pixels in the lower dimensional output as ``cells,'' where three $2\times2$ non-overlapping max pooling operations in the encoder result in $8\times8$ pixel cells. The encoder maps the input image $I\in \mathbb{R}^{H\times W}$ to an intermediate tensor $\mathcal{B}\in \mathbb{R}^{H_c \times W_c \times F}$ with smaller spatial dimension and greater channel depth (\ie, $H_c < H$, $W_c < W$ and $F > 1$).

\subsection{Interest Point Decoder}

For interest point detection, each pixel of the output corresponds to a probability of ``point-ness'' for that pixel in the input. The standard network design for dense prediction involves an encoder-decoder pair, where the spatial resolution is decreased via pooling or strided convolution, and then upsampled back to full resolution via upconvolution operations, such as done in SegNet~\cite{vijay15}. Unfortunately, upsampling layers tend to add a high amount of computation and can introduce unwanted checkerboard artifacts~\cite{odena2016deconvolution}, thus we designed the interest point detection head with an explicit decoder\footnote{This decoder has no parameters, and is known as ``sub-pixel convolution''~\cite{Shi2016} or ``depth to space'' in TensorFlow or ``pixel shuffle'' in PyTorch.} to reduce the computation of the model. 

The interest point detector head computes $\mathcal{X}\in \mathbb{R}^{H_c\times W_c \times 65}$ and outputs a tensor sized $\mathbb{R}^{H\times W}$. The $65$ channels correspond to local, non-overlapping $8\times 8$ grid regions of pixels plus an extra ``no interest point'' dustbin. After a channel-wise {\color{myblue}softmax}, the dustbin dimension is removed and a $\mathbb{R}^{H_c \times W_c \times 64} \Rightarrow\mathbb{R}^{H \times W}$ {\color{myblue}reshape} is performed.

\subsection{Descriptor Decoder}

\begin{figure}[t]
\begin{center}
\vspace{-.1in}
\includegraphics[width=\linewidth]{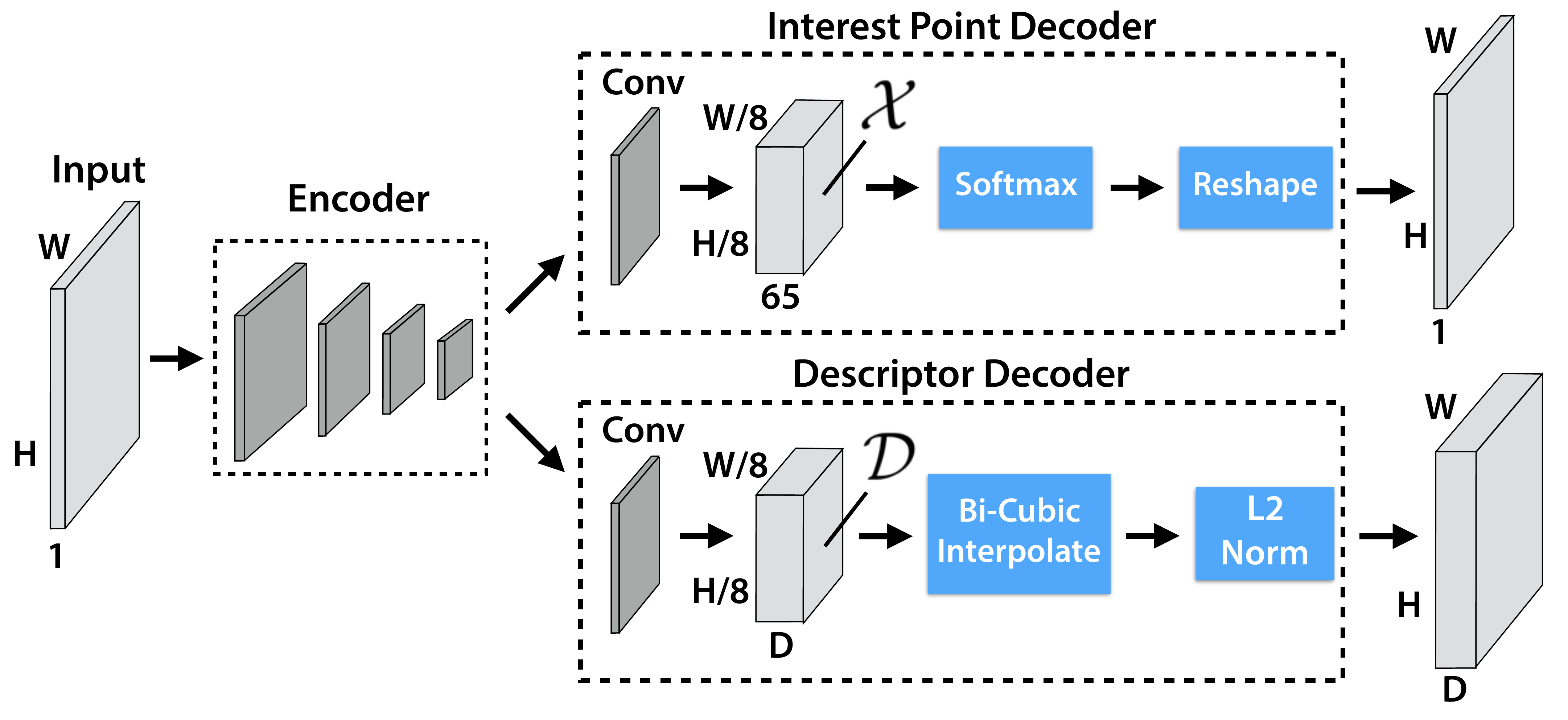}
\vspace{-.3in}
\end{center}
   \caption{\textbf{SuperPoint Decoders}. Both decoders operate on a shared and spatially reduced representation of the input. To keep the model fast and easy to train, both decoders use non-learned upsampling to bring the representation back to $\mathbb{R}^{H\times W}$.
 }
\label{fig:spnet}
\vspace{-.1in}
\end{figure}

The descriptor head computes $\mathcal{D} \in \mathbb{R}^{H_c\times W_c \times D}$ and outputs a tensor sized $\mathbb{R}^{H\times W \times D}$.
To output a dense map of L2-normalized fixed length descriptors, we use a model similar to UCN~\cite{choy_nips16} to first output a semi-dense grid of descriptors (\eg, one every 8 pixels). Learning descriptors semi-densely rather than densely reduces training memory and keeps the run-time tractable. The decoder then performs {\color{myblue}bi-cubic interpolation} of the descriptor and then {\color{myblue}L2-normalizes} the activations to be unit length. This fixed, non-learned descriptor decoder is shown in Figure~\ref{fig:spnet}.

\subsection{Loss Functions}
The final loss is the sum of two intermediate losses: one for the interest point detector, $\mathcal{L}_p$, and one for the descriptor, $\mathcal{L}_d$. We use pairs of synthetically warped images which have both (a) pseudo-ground truth interest point locations and (b) the ground truth correspondence from a randomly generated homography $\mathcal{H}$ which relates the two images. This allows us to optimize the two losses simultaneously, given a pair of images, as shown in Figure~\ref{fig:self_supervised}{\color{red}c}. We use $\lambda$ to balance the final loss:
\begin{align}
\begin{split}
\mathcal{L}(\mathcal{X}, \mathcal{X}' &, \mathcal{D}, \mathcal{D}'; Y, Y', S) = \\& \mathcal{L}_p(\mathcal{X},Y) + \mathcal{L}_p(\mathcal{X}',Y') + \lambda \mathcal{L}_d(\mathcal{D},\mathcal{D}',S). 
\end{split}
\end{align}

\begin{figure*}[t]
\begin{center}
\vspace{-.4in}
\includegraphics[width=0.98\linewidth]{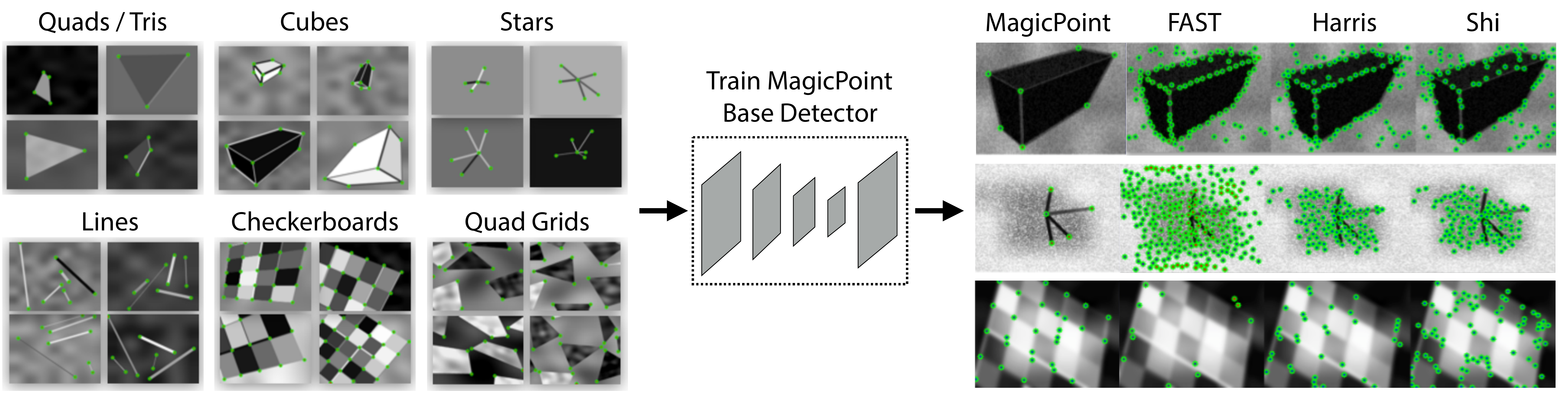}
\vspace{-.15in}
\end{center}
\caption{{\bf Synthetic Pre-Training.} We use our Synthetic Shapes dataset consisting of rendered triangles, quadrilaterals, lines, cubes, checkerboards, and stars each with ground truth corner locations. The dataset is used to train the MagicPoint convolutional neural network, which is more robust to noise when compared to classical detectors.} 
\label{fig:ss}
\end{figure*}

The interest point detector loss function $\mathcal{L}_p$ is a fully-convolutional cross-entropy loss over the cells ${\bf x}_{hw} \in \mathcal{X}$. We call the set of corresponding ground-truth interest point labels\footnote{If two ground truth corner positions land in the same bin then we randomly select one ground truth corner location.} $Y$ and individual entries as $y_{hw}$. The loss is: 
\vspace{-.05in}
\begin{equation}
  \mathcal{L}_p(\mathcal{X},Y) = \frac{1}{H_cW_c}\sum_{\substack{h=1 \\ w=1}}^{H_c, W_c}l_p({\bf x}_{hw}; y_{hw}),
  \label{eqn:loss-point}
\end{equation}
\vspace{-.1in}
where
\vspace{-.05in}
\begin{equation}
l_p({\bf x}_{hw};y) = -\log\left({\frac{\exp({\bf x}_{hwy})}{\sum_{k=1}^{65}\exp({\bf x}_{hwk})}}\right).
\end{equation}

The descriptor loss is applied to all pairs of descriptor cells, ${\bf d}_{hw} \in \mathcal{D}$ from the first image and ${\bf d'}_{h'w'} \in \mathcal{D}'$ from the second image. The homography-induced correspondence between the $(h,w)$ cell and the $(h',w')$ cell can be written as follows:
\vspace{-.05in}
\begin{equation}
    s_{hwh'w'}= 
\begin{cases}
    1,& \text{if } ||\widehat{\mathcal{H}{\bf p}_{hw}} - {\bf p}_{h'w'}|| \leq 8\\
    0,              & \text{otherwise}
\end{cases}
\end{equation}
where ${\bf p}_{hw}$ denotes the location of the center pixel in the $(h,w)$ cell, and $\widehat{\mathcal{H}{\bf p}_{hw}}$ denotes multiplying the cell location ${\bf p}_{hw}$ by the homography $\mathcal{H}$ and dividing by the last coordinate, as is usually done when transforming between Euclidean and homogeneous coordinates. We denote the entire set of correspondences for a pair of images with $S$.

We also add a weighting term $\lambda_d$ to help balance the fact that there are more negative correspondences than positive ones. We use a hinge loss with positive margin $m_p$ and negative margin $m_n$. The descriptor loss is defined as:
\vspace{-.05in}
\begin{align}
\begin{split}
\mathcal{L}_d(&\mathcal{D},\mathcal{D}',S) = \\& \frac{1}{(H_cW_c)^2}\sum_{\substack{h=1 \\ w=1}}^{H_c, W_c}\sum_{\substack{h'=1 \\ w'=1}}^{H_c, W_c} l_d({\bf d}_{hw},{\bf d}_{h'w'}'; s_{hwh'w'}),
\end{split}
\label{eqn:loss-descriptor}
\end{align}
\vspace{-.1in}
where
\vspace{-.1in}

\begin{equation}
\begin{split}
l_d({\bf d},{\bf d'}; s) = \lambda_d * s * \max(0, m_p - {\bf d}^T {\bf d}') \\ + (1-s) * \max(0, {\bf d}^T{\bf d}' - m_n).
\end{split}
\end{equation}

\section{Synthetic Pre-Training}
\label{sec-mp}
In this section, we describe our method for training a base detector (shown in Figure~\ref{fig:self_supervised}{\color{red}a}) called MagicPoint which is used in conjunction with Homographic Adaptation to generate pseudo-ground truth interest point labels for unlabeled images in a self-supervised fashion.

\subsection{Synthetic Shapes}
\label{sec:synthetic-shapes}

There is no large database of interest point labeled images that exists today. Thus to bootstrap our deep interest point detector, we first create a large-scale synthetic dataset called \emph{Synthetic Shapes} that consists of simplified 2D geometry via synthetic data rendering of quadrilaterals, triangles, lines and ellipses. Examples of these shapes are shown in Figure~\ref{fig:ss}. In this dataset, we are able to remove label ambiguity by modeling interest points with simple Y-junctions, L-junctions, T-junctions as well as centers of tiny ellipses and end points of line segments.

Once the synthetic images are rendered, we apply homographic warps to each image to augment the number of training examples. The data is generated on-the-fly and no example is seen by the network twice. While the types of interest points represented in Synthetic Shapes represents only a subset of all potential interest points found in the real world, we found it to work reasonably well in practice when used to train an interest point detector.

\begin{figure*}
\centering
\vspace{-.3in}
\includegraphics[width=.97\textwidth]{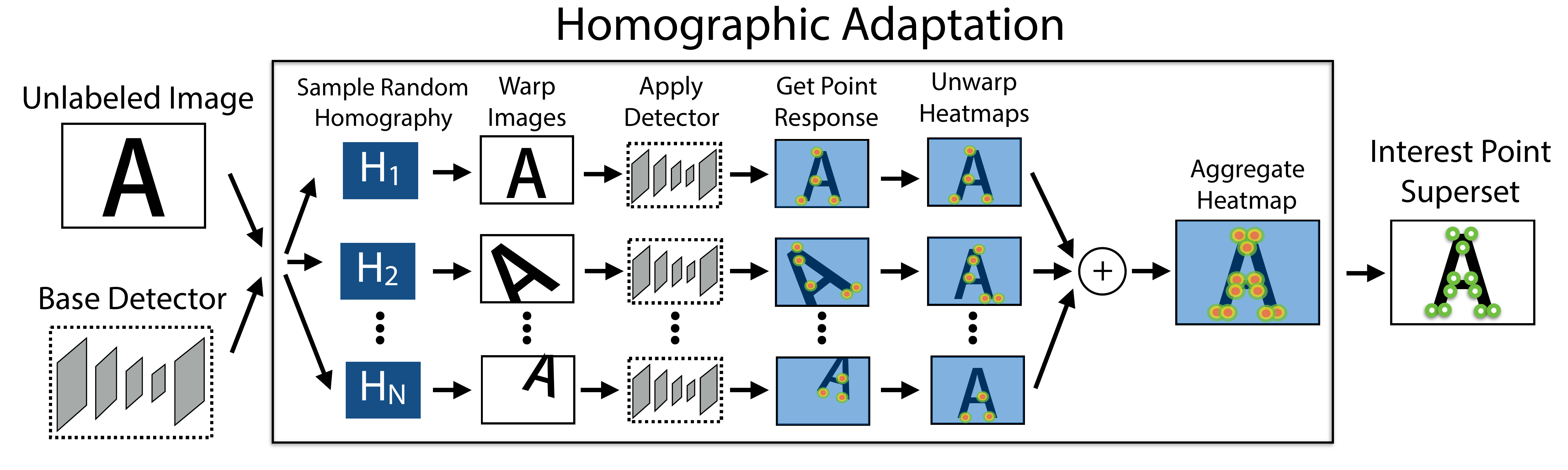}
\caption{{\bf Homographic Adaptation.} Homographic Adaptation is a form of self-supervision for boosting the geometric consistency of an interest point detector trained with convolutional neural networks. The entire procedure is mathematically defined in Equation~\ref{eqn:ha}. \label{fig:ha_train}} 
\end{figure*}

\subsection{MagicPoint}
We use the detector pathway of the SuperPoint architecture (ignoring the descriptor head) and train it on Synthetic Shapes. We call the resulting model \emph{MagicPoint}.

Interestingly, when we evaluate MagicPoint against other traditional corner detection approaches such as FAST~\cite{rosten2006}, Harris corners~\cite{harris1988} and Shi-Tomasi's ``Good Features To Track''~\cite{shi1994} on the Synthetic Shapes dataset, we discovered a large performance gap in our favor. We measure the mean Average Precision (mAP) on $1000$ held-out images of the Synthetic Shapes dataset, and report the results in Table~\ref{tbl:ss}. The classical detectors struggle in the presence of imaging noise -- qualitative examples of this are shown in Figure~\ref{fig:ss}. More detailed experiments can be found in Appendix~\ref{sec:extra-ss}.

\begin{table}[h]
  \centering
  \small
  \def\arraystretch{1.1}
  \setlength{\tabcolsep}{2pt}
  \begin{tabular}{cccccc}
          \toprule
              & & MagicPoint     & FAST  & Harris & Shi \\
          \midrule
          mAP & no noise &   \textbf{0.979} & 0.405 & 0.678  & 0.686  \\
          mAP & noise & \textbf{0.971} & 0.061 & 0.213  & 0.157  \\
          \bottomrule
  \end{tabular}
  \bigskip
  \caption{\textbf{Synthetic Shapes Detector Performance}. The MagicPoint model outperforms classical detectors in detecting corners of simple geometric shapes and is robust to added noise.}
  \label{tbl:ss}
\end{table}

The MagicPoint detector performs very well on Synthetic Shapes, but does it generalize to real images? To summarize a result that we later present in Section~\ref{hpatches_rep}, the answer is \emph{yes}, but not as well as we hoped. We were surprised to find that MagicPoint performs reasonably well on real world images, especially on scenes which have strong corner-like structure such as tables, chairs and windows. Unfortunately in the space of all natural images, it under-performs when compared to the same classical detectors on repeatability under viewpoint changes. This motivated our self-supervised approach for training on real-world images which we call Homographic Adaptation.

\section{Homographic Adaptation}
\label{sec-ha}

Our system bootstraps itself from a base interest point detector and a large set of unlabeled images from the target domain (\eg, MS-COCO). Operating in a self-supervised paradigm (also known as self-training), we first generate a set of pseudo-ground truth interest point locations for each image in the target domain, then use traditional supervised learning machinery. At the core of our method is a process that applies random homographies to warped copies of the input image and combines the results -- a process we call \emph{Homographic Adaptation} (see Figure~\ref{fig:ha_train}).

\subsection{Formulation}
Homographies give exact or almost exact image-to-image transformations for camera motion with only rotation around the camera center, scenes with large distances to objects, and planar scenes. Moreover, because most of the world is reasonably planar, a homography is good model for what happens when the same 3D point is seen from different viewpoints. Because homographies do not require 3D information, they can be randomly sampled and easily applied to any 2D image -- involving little more than bilinear interpolation. For these reasons, \emph{homographies are at the core of our self-supervised approach}.

Let $f_{\theta}(\cdot)$ represent the initial interest point function we wish to adapt, $I$ the input image, ${\bf x}$ the resulting interest points and $\mathcal{H}$ a random homography, so that:

\vspace{-.1in}
\begin{equation}
  {\bf x} = f_\theta(I).
\end{equation}

An ideal interest point operator should be covariant with respect to homographies. A function $f_\theta(\cdot)$ is covariant with $\mathcal{H}$ if the output transforms with the input. In other words, a covariant detector will satisfy, for all $\mathcal{H}$ \footnote{For clarity, we slightly abuse notation and allow $\mathcal{H}{\bf x}$ to denote the homography matrix $\mathcal{H}$ being applied to the resulting interest points, and $\mathcal{H}(I)$ to denote the entire image $I$ being warped by $\mathcal{H}$.}:

\vspace{-.1in}
\begin{equation}
  \mathcal{H}{\bf x}=f_\theta(\mathcal{H}(I)),
\end{equation}
moving homography-related terms to the right, we get:
\begin{equation}
  {\bf x} = \mathcal{H}^{-1}f_\theta(\mathcal{H}(I)).
  \label{eqn:eq1}
\end{equation}
In practice, a detector will not be perfectly covariant -- different homographies in Equation~\ref{eqn:eq1} will result in different interest points ${\bf x}$. The basic idea behind Homographic Adaptation is to perform an empirical sum over a sufficiently large sample of random $\mathcal{H}$'s (see Figure~\ref{fig:ha_train}). The resulting aggregation over samples thus gives rise to a new and improved, super-point detector,
${\hat F}(\cdot)$:

\vspace{-.1in}
\begin{equation}
  {\hat F}(I;f_\theta) = \frac{1}{N_h}\sum_{i=1}^{N_h} \mathcal{H}_i^{-1}f_\theta(\mathcal{H}_i(I)).
    \label{eqn:ha}
\end{equation}

\begin{figure}
\centering
\includegraphics[width=\linewidth]{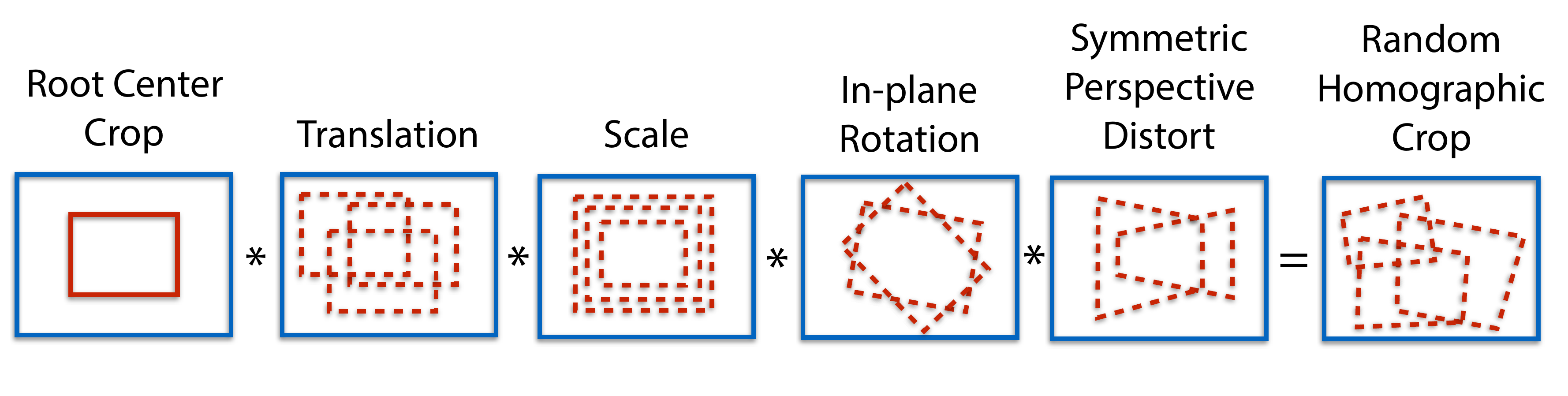}
\caption{{\bf Random Homography Generation.} We generate random homographies as the composition of less expressive, simple transformations.}
\label{fig:choosing_h}
\end{figure}

\subsection{Choosing Homographies}
\label{sec:choose_h}
Not all 3x3 matrices are good choices for Homographic Adaptation. To sample good homographies which represent plausible camera transformations, we decompose a potential homography into more simple, less expressive transformation classes. We sample within pre-determined ranges for translation, scale, in-plane rotation, and symmetric perspective distortion using a truncated normal distribution. These transformations are composed together with an initial root center crop to help avoid bordering artifacts. This process is shown in Figure~\ref{fig:choosing_h}.

When applying Homographic Adaptation to an image, we use the average response across a large number of homographic warps of the input image. The number of homographic warps $N_h$ is a hyper-parameter of our approach. We typically enforce the first homography to be equal to identity, so that $N_h$=1 in our experiments corresponds to doing no adaptation. We performed an experiment to determine the best value for $N_h$, varying the range of $N_h$ from “small” $N_h=10$, to “medium” $N_h=100$, and “large” $N_h=1000$. Our experiments suggest that there is diminishing returns when performing more than $100$ homographies. On a held-out set of images from MS-COCO, we obtain a repeatability score of $.67$ without any Homographic Adaptation, a repeatability boost of 21\% when performing $N_h=100$ transforms, and a repeatability boost of 22\% when $N_h=1000$, thus the added benefit of using more than $100$ homographies is minimal. For a more detailed analysis and discussion of this experiment see Appendix~\ref{sec:extra-ha}.

\subsection{Iterative Homographic Adaptation}
\begin{figure}[t]
\begin{center}
\includegraphics[width=\linewidth]{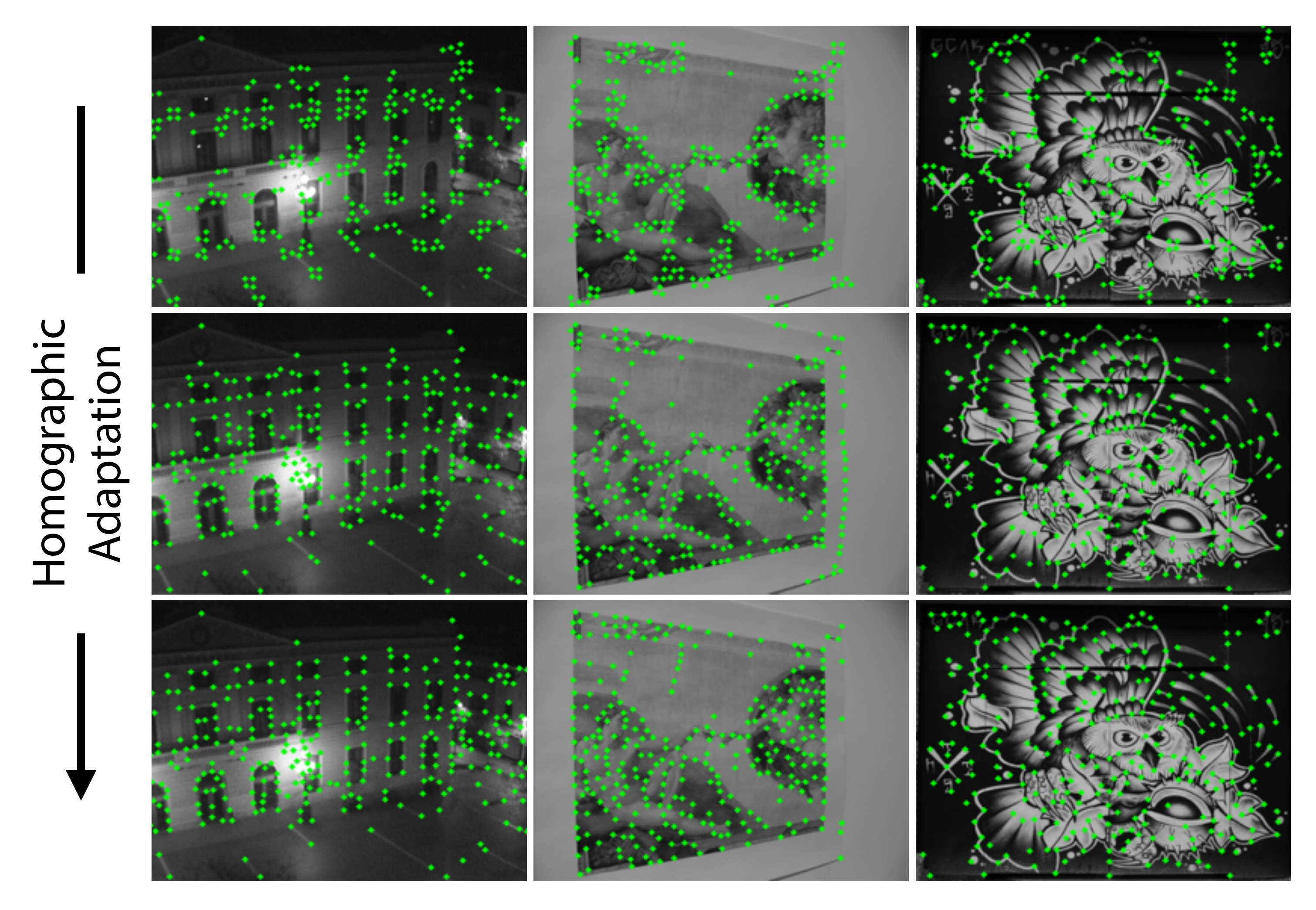}
\end{center}
   \caption{\textbf{Iterative Homographic Adaptation}. Top row: initial base detector (MagicPoint) struggles to find repeatable detections. Middle and bottom rows: further training with Homographic Adaption improves detector performance.}
\label{fig:progression}
\vspace{-.1in}
\end{figure}

We apply the Homographic Adaptation technique at training time to improve the generalization ability of the base MagicPoint architecture on real images. The process can be repeated iteratively to continually self-supervise and improve the interest point detector. In all of our experiments, we call the resulting model, after applying Homographic Adaptation, \emph{SuperPoint} and show the qualitative progression on images from HPatches in Figure~\ref{fig:progression}.

\section{Experimental Details}
\label{sec:exp-details}

In this section we provide some implementation details for training the MagicPoint and SuperPoint models.  This encoder has a VGG-like~\cite{vgg} architecture that has eight 3x3 convolution layers sized 64-64-64-64-128-128-128-128. Every two layers there is a 2x2 max pool layer. Each decoder head has a single 3x3 convolutional layer of 256 units followed by a 1x1 convolution layer with 65 units and 256 units for the interest point detector and descriptor respectively. All convolution layers in the network are followed by ReLU non-linear activation and BatchNorm normalization.

To train the fully-convolutional SuperPoint model, we start with a base MagicPoint model trained on Synthetic Shapes. The MagicPoint architecture is the SuperPoint architecture without the descriptor head. The MagicPoint model is trained for 200,000 iterations of synthetic data. Since the synthetic data is simple and fast to render, the data is rendered on-the-fly, thus no single example is seen twice by the network.

We generate pseudo-ground truth labels using the MS-COCO 2014~\cite{coco} training dataset split which has 80,000 images and the MagicPoint base detector. The images are sized to a resolution of $240\times 320$ and converted to grayscale. The labels are generated using Homographic Adaptation with $N_h=100$, as motivated by our results from Section~\ref{sec:choose_h}. We repeat the Homographic Adaptation a second time, using the resulting model trained from the first round of Homographic Adaptation.

The joint training of SuperPoint is also done on $240 \times 320$ grayscale COCO images. For each training example, a homography is randomly sampled. It is sampled from a more restrictive set of homographies than during Homographic Adaptation to better model the target application of pair-wise matching (\eg, we avoid sampling extreme in-plane rotations as they are rarely seen in HPatches). The image and corresponding pseudo-ground truth are transformed by the homography to create the needed inputs and labels. The descriptor size used in all experiments is $D=256$. We use a weighting term of $\lambda_d=250$ to keep the descriptor learning balanced. The descriptor hinge loss uses a positive margin $m_p=1$ and negative margin $m_n=0.2$. We use a factor of $\lambda=0.0001$ to balance the two losses.

All training is done using PyTorch~\cite{pytorch} with mini-batch sizes of 32 and the ADAM solver with default parameters of $lr=0.001$ and $\beta=(0.9, 0.999)$. We also use standard data augmentation techniques such as random Gaussian noise, motion blur, brightness level changes to improve the network's robustness to lighting and viewpoint changes.
\newpage

\section{Experiments}
\label{eval}
\vspace{.05in}
In this section we present quantitative results of the methods presented in the paper. Evaluation of interest points and descriptors is a well-studied topic, thus we follow the evaluation protocol of Miko\l{}ajczyk \etal~\cite{Mikolajczyk2005}. For more details on our evaluation metrics, see Appendix~\ref{sec:metrics}.

\subsection{System Runtime}
\vspace{.05in}
We measure the run-time of the SuperPoint architecture using a Titan X GPU and the timing tool that comes with the Caffe~\cite{caffe} deep learning library. A single forward pass of the model runs in approximately $11.15$ ms with inputs sized $480 \times 640$, which produces the point detection locations and a semi-dense descriptor map. To sample the descriptors at the higher $480 \times 640$ resolution from the semi-dense descriptor, it is not necessary to create the entire dense descriptor map -- we can just sample from the 1000 detected locations, which takes about $1.5$ ms on a CPU implementation of bi-cubic interpolation followed by L2 normalization. Thus we estimate the total runtime of the system on a GPU to be about $13$ ms or $\mathbf{70}$ \textbf{FPS}.  

\subsection{HPatches Repeatability}
\vspace{.05in}
\label{hpatches_rep}
In our experiments we train SuperPoint on the MS-COCO images, and evaluate using the HPatches dataset~\cite{hpatches17}. HPatches contains 116 scenes with 696 unique images. The first 57 scenes exhibit large changes in illumination and the other 59 scenes have large viewpoint changes.


\begin{table}[t]
  \centering
  \scriptsize
  \def\arraystretch{1.1}
  \setlength{\tabcolsep}{2pt}
  \begin{tabular}{cccccccccc}
          \toprule
          & \multicolumn{2}{c}{57 Illumination Scenes} & \multicolumn{2}{c}{59 Viewpoint Scenes} \\
          & NMS=4 & NMS=8 & NMS=4 & NMS=8 \\
          \midrule
          {\it SuperPoint} & \bf{.652} & \bf{.631} & .503          & \textbf{.484}  \\
          {\it MagicPoint} & .575      & .507      & .322          & .260  \\
          {\it FAST}       & .575      & .472      & .503          & .404  \\
          {\it Harris}     & .620      & .533      & \textbf{.556} & .461 \\
          {\it Shi}        & .606      & .511      & .552          & .453 \\
          {\it Random}     & .101      & .103      & .100          & .104 \\
          \bottomrule
  \end{tabular}
  \bigskip
  \caption{\textbf{HPatches Detector Repeatability}. SuperPoint is the most repeatable under illumination changes, competitive on viewpoint changes, and outperforms MagicPoint in all scenarios. }
  \label{tbl:detector}
\end{table}

\begin{figure*}
\centering
\includegraphics[width=\textwidth]{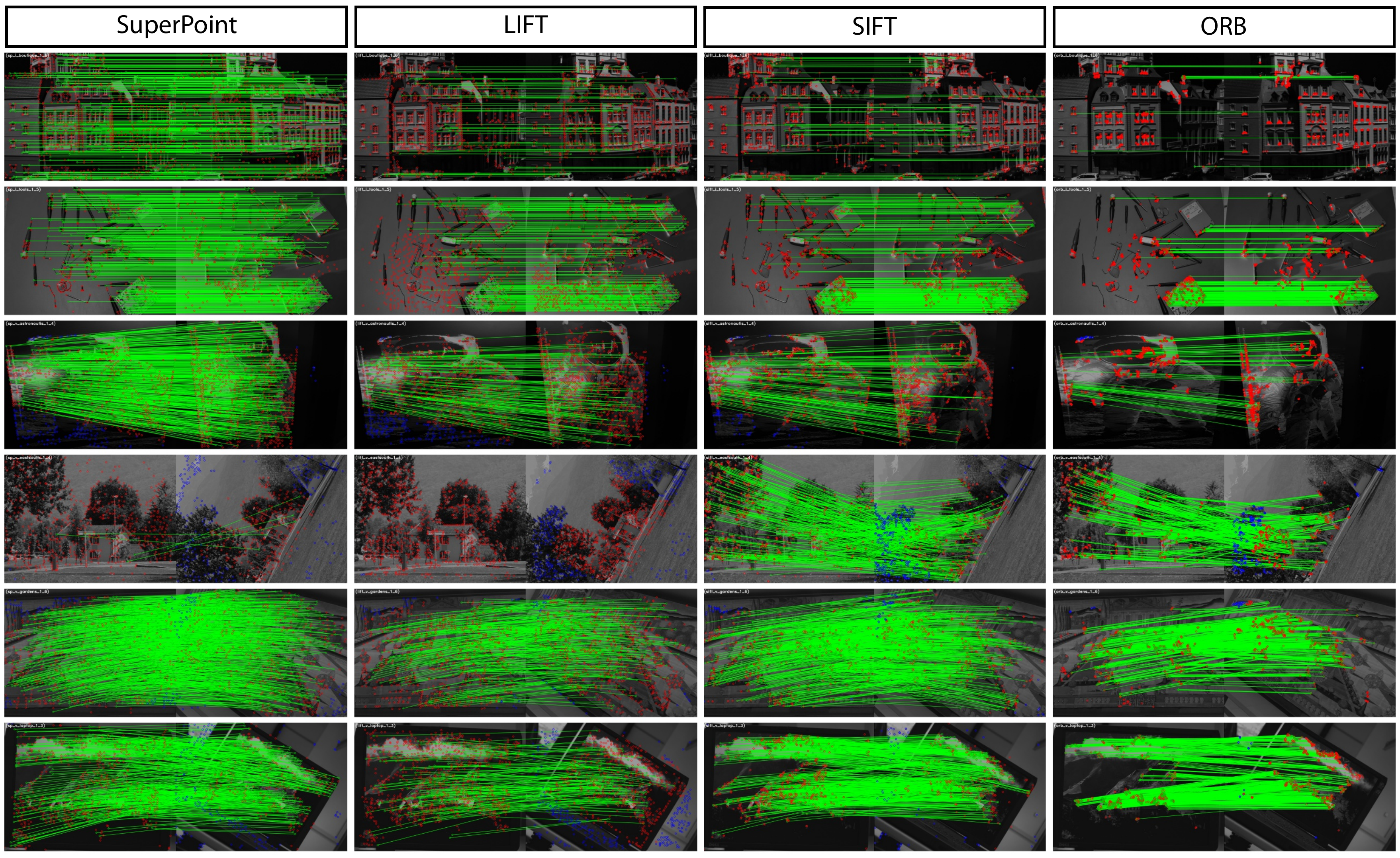}
\caption{{\bf Qualitative Results on HPatches}. The green lines show correct correspondences. SuperPoint tends to produce more dense and correct matches compared to LIFT, SIFT and ORB. While ORB has the highest average repeatability, the detections cluster together and generally do not result in more matches or more accurate homography estimates (see \ref{tbl:hest}). Row 4: Failure case of SuperPoint and LIFT due to extreme in-plane rotation not seen in the training examples. See Appendix \ref{sec:extra-dumps} for additional homography estimation example pairs.}
\label{h_qualitative}
\end{figure*}

To evaluate the interest point detection ability of the SuperPoint model, we measure repeatability on the HPatches dataset. We compare it to the MagicPoint model (before Homographic Adaptation), as well as FAST~\cite{rosten2006}, Harris~\cite{harris1988} and Shi~\cite{shi1994}, all implemented using OpenCV. Repeatability is computed at $240 \times 320$ resolution with $300$ points detected in each image. We also vary the Non-Maximum Suppression (NMS) applied to the detections. We use a correct distance of $\epsilon = 3$ pixels. Applying larger amounts of NMS helps ensure that the points are evenly distributed in the image, useful for certain applications such as ORB-SLAM~\cite{mur2015}, where a minimum number of FAST corner detections is forced in each cell of a coarse grid. 

In summary, the Homographic Adaptation technique used to transform MagicPoint into SuperPoint gives a large boost in repeatability, especially under large viewpoint changes. Results are shown in Table~\ref{tbl:detector}. The SuperPoint model outperforms classical detectors under illumination changes and performs on par with classical detectors under viewpoint changes.

\subsection{HPatches Homography Estimation}
To evaluate the performance of the SuperPoint interest point detector and descriptor network, we compare matching ability on the HPatches dataset. We evaluate SuperPoint against three well-known detector and descriptor systems: LIFT~\cite{yi16}, SIFT~\cite{lowe2004} and ORB~\cite{rublee2011}. For LIFT we use the pre-trained model (Picadilly) provided by the authors. For SIFT and ORB we use the default OpenCV implementations. We use a correct distance of $\epsilon = 3$ pixels for Rep, MLE, NN mAP and MScore. We compute a maximum of $1000$ points for all systems at a $480 \times 640$ resolution and compute a number of metrics for each image pair. To estimate the homography, we perform nearest neighbor matching from all interest points+descriptors detected in the first image to all the interest points+descriptors in the second. We use an OpenCV implementation (\texttt{findHomography()} with RANSAC) with all the matches to compute the final homography estimate.

\begin{table}[t]
  \centering
  \scriptsize
  \def\arraystretch{1.1}
  \setlength{\tabcolsep}{2pt}
  \begin{tabular}{cccc|cccc}
          \toprule
          & \multicolumn{3}{c|}{Homography Estimation} & \multicolumn{2}{c}{Detector Metrics} & \multicolumn{2}{c}{Descriptor Metrics} \\
          \midrule
          & $\epsilon = 1$ & $\epsilon = 3$ & $\epsilon = 5$ & Rep. & MLE & NN mAP & M. Score  \\
          \midrule
          {\it SuperPoint} & .310      & \bf{.684} & \bf{.829} & .581  & 1.158  & \bf{.821} & \bf{.470} \\
          {\it LIFT}       & .284      & .598      & .717      & .449  & 1.102  & .664    & .315        \\
          {\it SIFT}       & \bf{.424} & .676      & .759      & .495  & \bf{0.833}  & .694    & .313        \\
          {\it ORB}        & .150      & .395      & .538      & \bf{.641}  & 1.157  &  .735  & .266       \\
          \bottomrule
  \end{tabular}
  \bigskip
  \caption{\textbf{HPatches Homography Estimation.} SuperPoint outperforms LIFT and ORB and performs comparably to SIFT using various $\epsilon$ thresholds of correctness. We also report related metrics which measure detector and descriptor performance individually. }
  \label{tbl:hest}
\end{table}

The homography estimation results are shown in Table~\ref{tbl:hest}. SuperPoint outperforms LIFT and ORB and performs comparably to SIFT for homography estimation on HPatches using various $\epsilon$ thresholds of correctness. Qualitative examples of SuperPoint versus LIFT, SIFT and ORB are shown in Figure~\ref{h_qualitative}. Please see Appendix \ref{sec:extra-dumps} for even more homography estimation example pairs. SuperPoint tends to produce a larger number of correct matches which densely cover the image, and is especially effective against illumination changes.

Quantitatively we outperform LIFT in almost all metrics. LIFT is also outperformed by SIFT in most metrics. This may be due to the fact that HPatches includes indoor sequences and LIFT was trained on a single outdoor sequence. Our method was trained on hundreds of thousands of warped MS-COCO images that exhibit a much larger diversity and more closely match the diversity in HPatches.

SIFT performs well for sub-pixel precision homographies $\epsilon=1$ and has the lowest mean localization error (MLE). This is likely due to the fact that SIFT performs extra sub-pixel localization, while other methods do not perform this step.

ORB achieves the highest repeatability (Rep.); however, its detections tend to form sparse clusters throughout the image as shown in Figure~\ref{h_qualitative}, thus scoring poorly on the final homography estimation task. This suggests that \emph{optimizing solely for repeatability does not result in better matching or estimation further up the pipeline}. 

SuperPoint scores strongly in descriptor-focused metrics such as nearest neighbor mAP (NN mAP) and matching score (M. Score), which confirms findings from both Choy \etal~\cite{choy_nips16} and Yi \etal~\cite{yi16} which show that learned representations for descriptor matching outperform hand-tuned representations. 

\vspace{-.05in}
\section{Conclusion}
\vspace{-.05in}
\label{sec-conclusion}
We have presented a fully-convolutional neural network architecture for interest point detection and description trained using a self-supervised domain adaptation framework called Homographic Adaptation. Our experiments demonstrate that $(1)$ it is possible to transfer knowledge from a synthetic dataset onto real-world images, $(2)$ sparse interest point detection and description can be cast as a single, efficient convolutional neural network, and $(3)$ the resulting system works well for geometric computer vision matching tasks such as Homography Estimation.

Future work will investigate whether Homographic Adaptation can boost the performance of models such as those used in semantic segmentation (\eg, SegNet~\cite{vijay15} ) and object detection (\eg, SSD~\cite{LiuAESR15}). It will also carefully investigate the ways that interest point detection and description (and potentially other tasks) benefit each other.

Lastly, we believe that our SuperPoint network can be used to tackle all visual data-association in 3D computer vision problems like SLAM and SfM, and that a learning-based Visual SLAM front-end will enable more robust applications in robotics and augmented reality.

\clearpage
\bibliographystyle{ieee}
\bibliography{main}

\appendix
\clearpage

\noindent{\huge\bfseries APPENDIX\par}

\section{Evaluation Metrics}
\label{sec:metrics}

In this section we present more details on the metrics used for evaluation. In our experiments we follow the protocol of~\cite{Mikolajczyk2005}, with one exception. Since our fully-convolutional model does not use local patches, we instead compare detection distances by measuring the distance between the 2D detection centers, rather than measure patch overlap. For multi-scale methods such as SIFT and ORB, we compare distances at the highest resolution scale.

\textbf{Corner Detection Average Precision.} We compute Precision-Recall curves and the corresponding Area-Under-Curve (also known as Average Precision), the pixel location error for correct detections, and the repeatability rate. For corner detection, we use a threshold $\varepsilon$ to determine if a returned point location $\mathbf{x}$ is correct relative to a set of $K$ ground-truth corners $\{\mathbf{ {\hat x}}_1, \dots, \mathbf{ {\hat x}}_K \}$. We define the correctness as follows:
\begin{equation}
  \verb+Corr+(\mathbf{x}) = (\min_{j} ||\mathbf{x} - {\bf {\hat x}}_j||) \leq \varepsilon.
  \label{eqn:correct-detection}
\end{equation}
The precision recall curve is created by varying the detection
confidence and summarized with a single number, namely the Average Precision (which ranges from $0$ to $1$), and larger AP is better.

\textbf{Localization Error.} To complement the AP analysis, we compute the corner localization error, but solely for the correct detections. We define the Localization Error as follows:
\begin{equation}
  \verb+LE+ = \frac{1}{N} \sum_{i : \verb+Corr+(\mathbf{x}_i)}
  \min_{j \in \{1,\dots,K\}}||\mathbf{x}_i - \mathbf{ {\hat x}}_j||.
  \label{eqn:mse}
\end{equation}
The Localization Error is between $0$ and $\varepsilon$, and lower LE is better.

\textbf{Repeatability.} 
We compute the repeatability rate for an interest point detector on a pair of images. Since the SuperPoint architecture is fully-convolutional and does not rely on patch extraction, we cannot compute patch overlap and instead compute repeatability by measuring the distance between the extracted 2D point centers. We use $\varepsilon$ to represent the correct distance threshold between two points. More concretely, let us assume we have $N_1$ points in the first image and $N_2$ points in the second image. We define correctness for repeatability experiments as follows:

\begin{equation}
  \verb+Corr+(\mathbf{x}_i) = (\min_{j \in \{1,\dots,N_2\}} ||\mathbf{x}_i - \mathbf{{\hat x}}_j||) \leq \varepsilon.
  \label{eqn:repeatability-detection}
\end{equation}
Repeatability simply measures the probability that a point is detected in the second image.
\begin{equation}
  \verb+Rep+ = \frac{1}{N_1+N_2} (\sum_{i} \verb+Corr+({\bf x}_i) + \sum_{j} \verb+Corr+({\bf x}_j)).
\end{equation}

\textbf{Nearest Neighbor mean Average Precision}. This metric captures how discriminating the descriptor is by evaluating it at multiple descriptor distance thresholds. It is computed by measuring Area Under Curve (AUC) of the Precision-Recall curve, using the Nearest Neighbor matching strategy. This metric is computed symmetrically across the pair of images and averaged. 

\textbf{Matching Score}. This metric measures the overall performance of the interest point detector and descriptor combined. It measures the ratio of ground truth correspondences that can be recovered by the whole pipeline over the number of features proposed by the pipeline in the shared viewpoint region. This metric is computed symmetrically across the pair of images and averaged.

\textbf{Homography Estimation.} We measure the ability of an algorithm to estimate the homography relating a pair of images by comparing the estimated homography $\hat{\mathcal{H}}$ to the ground truth homography $\mathcal{H}$. It is not straightforward to compare the $3\times 3$ $\mathcal{H}$ matrices directly, since different entries in the matrix have different scales. Instead we compare how well the homography transforms the four corners of one image onto the other. We define the four corners of the first image as $\mathbf{c}_1, \mathbf{c}_2, \mathbf{c}_3, \mathbf{c}_4$. We then apply the ground truth $\mathcal{H}$ to get the ground truth corners in the second image $\mathbf{c}'_1, \mathbf{c}'_2, \mathbf{c}'_3, \mathbf{c}'_4$ and the estimated homography $\hat{\mathcal{H}}$ to get $\mathbf{\hat{c}}'_1, \mathbf{\hat{c}}'_2, \mathbf{\hat{c}}'_3, \mathbf{\hat{c}}'_4$. We use a threshold $\varepsilon$ to denote a correct homography.
\begin{equation}
  \verb+CorrH+ = \frac{1}{N} \sum_{i=1}^{N}
  \left( \left( \frac{1}{4} \sum_{j=1}^{4}||\mathbf{ c'}_{ij} - \mathbf{ {\hat c'}}_{ij}|| \right) \leq \varepsilon \right).
\end{equation}
The scores range between $0$ and $1$, higher is better.

\section{Additional Synthetic Shapes Experiments} 
\label{sec:extra-ss}

We present the full results of the SuperPoint interest point detector (ignoring the descriptor head) trained and evaluated on the Synthetic Shapes dataset.\footnote{An earlier version of our MagicPoint experiments can be found in our ``Toward Geometric DeepSLAM'' paper~\cite{detone17a}.}  We call this detector \emph{MagicPoint}. The data consists of simple synthetic geometry that a human could easily label with the ground truth corner locations. We expect a good point detector to easily detect the correct corners in these scenarios. In fact, we were surprised at how difficult the simple geometries were for the classical point detectors such as FAST~\cite{rosten2006}, Harris~\cite{harris1988} and the Shi-Tomasi ``Good Features to Track''~\cite{shi1994}.

We evaluated two models: \emph{MagicPointL} and \emph{MagicPointS}. Both models share the same encoder architecture, but differ in the number of neurons per layer. MagicPointL and MagicPointS have 64-64-64-64-128-128-128-128-128 and 9-9-16-16-32-32-32-32-32 respectively.

\begin{figure}[h]
\centering
\vspace{-.1in}
\includegraphics[width=\linewidth]{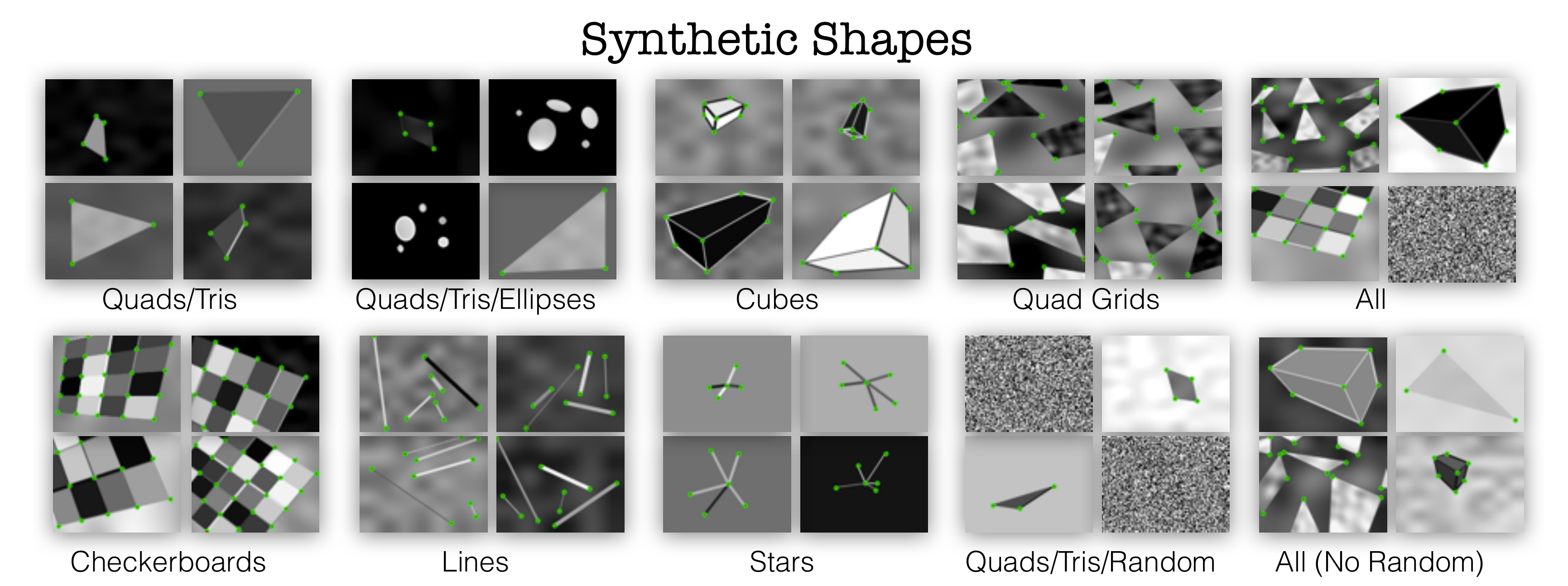}
\caption{{\bf Synthetic Shapes Dataset.} The Synthetic Shapes dataset consists of rendered triangles, quadrilaterals, lines, cubes, checkerboards, and stars each with ground truth corner locations. It also includes some negative images with no ground truth corners, such as ellipses and random noise images. \label{fig:mpdata}} 
\end{figure}

We created an evaluation dataset with our Synthetic Shapes generator to determine how well our detector is able to localize simple corners. There are 10 categories of images, shown in Figure~\ref{fig:mpdata}.

\begin{figure}
\centering
\includegraphics[width=\linewidth]{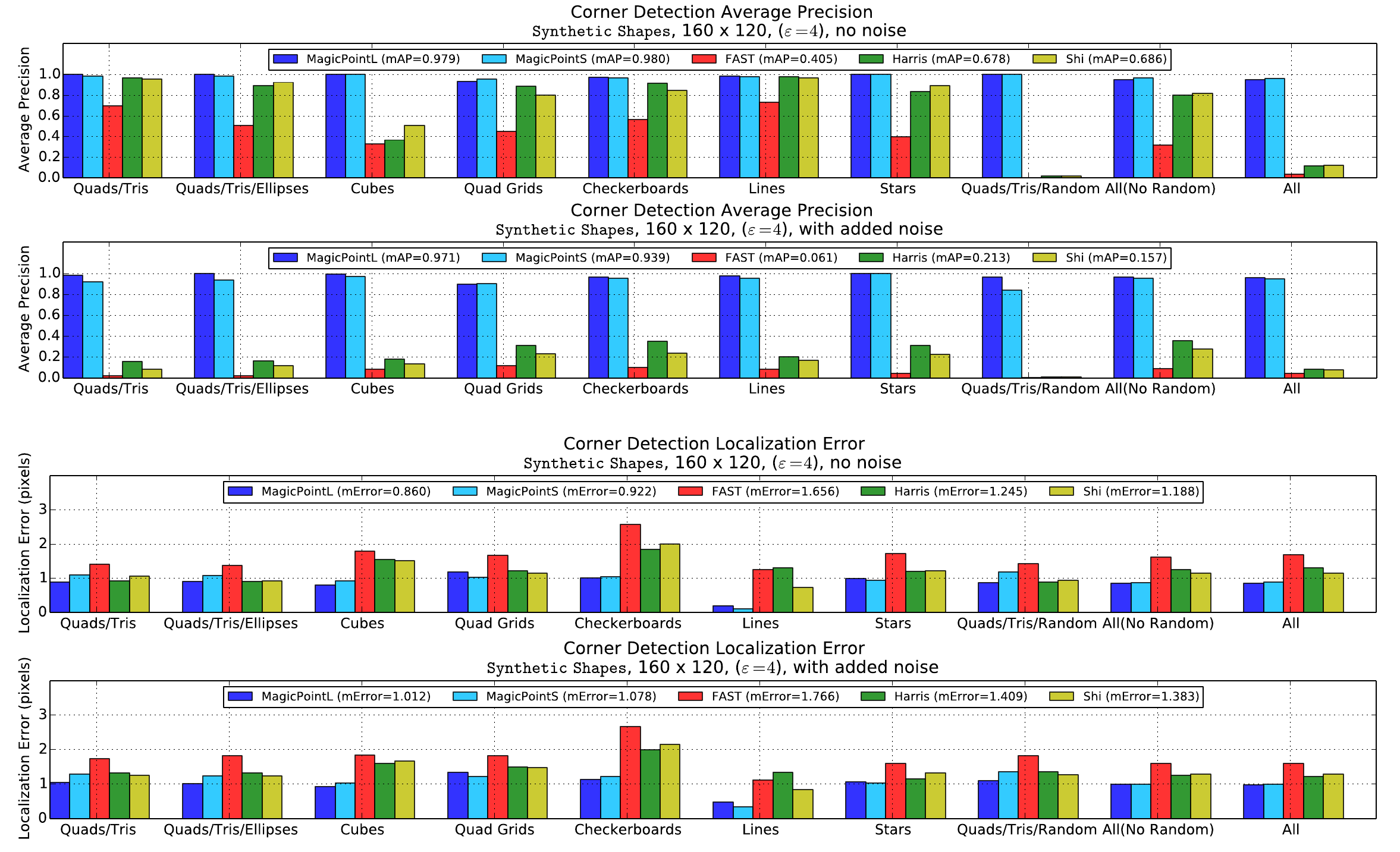}
\caption{{\bf Per Shape Category Results.} These plots report Average Precision and Corner Localization Error for each of the 10 categories in the Synthetic Shapes dataset with and without noise. The sequences with ``Random'' inputs are especially difficult for the classical detectors. \label{fig:mpeval}} 
\vspace{-.1in}
\end{figure}

\textbf{Mean Average Precision and Mean Localization Error.} For each category, there are 1000 images sampled from the Synthetic Shapes generator. We compute Average Precision and Localization Error with and without added imaging noise. A summary of the per category results are shown in Figure~\ref{fig:mpeval} and the mean results are shown in Table~\ref{table:mptable_ss}. The MagicPoint detectors outperform the classical detectors in all categories. There is a significant performance gap in mAP in all categories in the presence of noise.

\begin{table}[t]
  \centering
  \scriptsize
  \def\arraystretch{1.1}
  \setlength{\tabcolsep}{2pt}
  \begin{tabular}{llccccccccc}
    \toprule
     Metric & Noise & MagicPointL & MagicPointS & FAST & Harris & Shi  \\
     \midrule
     mAP & no noise & 0.979 & \textbf{0.980} & 0.405 & 0.678 & 0.686 \\ 
     mAP & noise & \textbf{0.971} & 0.939 & 0.061 & 0.213 & 0.157 \\
     MLE & no noise & \textbf{0.860} & 0.922 & 1.656 & 1.245 & 1.188 \\
     MLE & noise & \textbf{1.012} & 1.078 & 1.766 & 1.409 & 1.383 \\
     \bottomrule
  \end{tabular}
  \bigskip
  \vspace{-.1in}
  \caption{\textbf{Synthetic Shapes Results Table.} Reports the mean Average Precision (mAP, higher is better) and Mean Localization Error (MLE, lower is better) across the 10 categories of images on the Synthetic Shapes dataset. Note that MagicPointL and MagicPointS are relatively unaffected by imaging noise.}
  \label{table:mptable_ss}
\end{table}

\begin{figure}
\centering
\includegraphics[width=\linewidth]{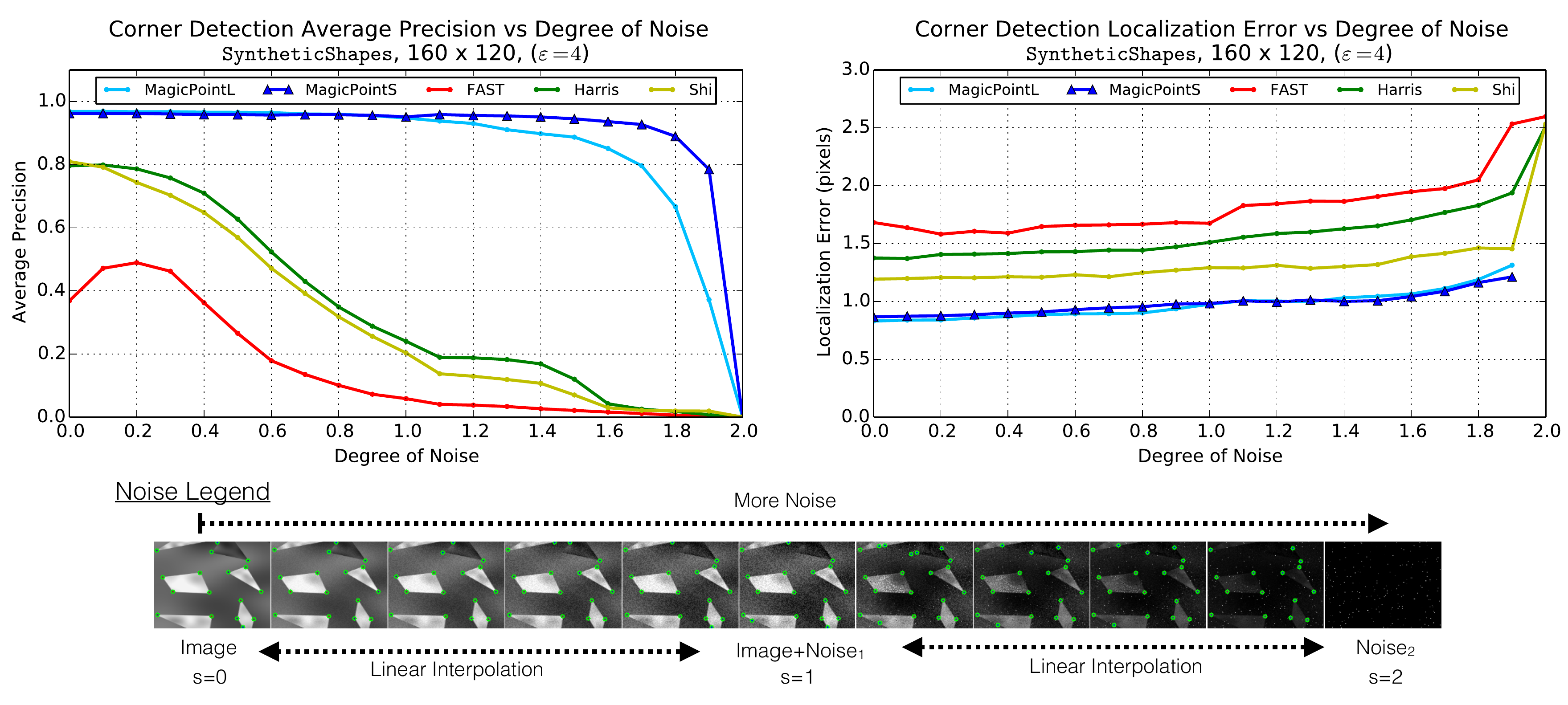}
\caption{{\bf Effect of Noise Magnitude.} Two versions of MagicPoint are compared to three classical point detectors on the Synthetic Shapes dataset (shown in Figure~\ref{fig:mpdata}). The MagicPoint models outperform the classical techniques in both metrics, especially in the presence of image noise. \label{fig:mpeval2}} 
\end{figure}

\textbf{Effect of Noise Magnitude.} Next we study the effect of noise more carefully by varying its magnitude. We were curious if the noise we add to the images is too extreme and unreasonable for a point detector. To test this hypothesis, we linearly interpolate between the clean image ($s=0$) and the noisy image ($s=1$). To push the detectors to the extreme, we also interpolate between the noisy image and random noise ($s=2$). The random noise images contain no geometric shapes, and thus produce an mAP score of $0.0$ for all detectors. An example of the varying degree of noise and the plots are shown in Figure~\ref{fig:mpeval2}.

\begin{figure}
\centering
\includegraphics[width=\linewidth]{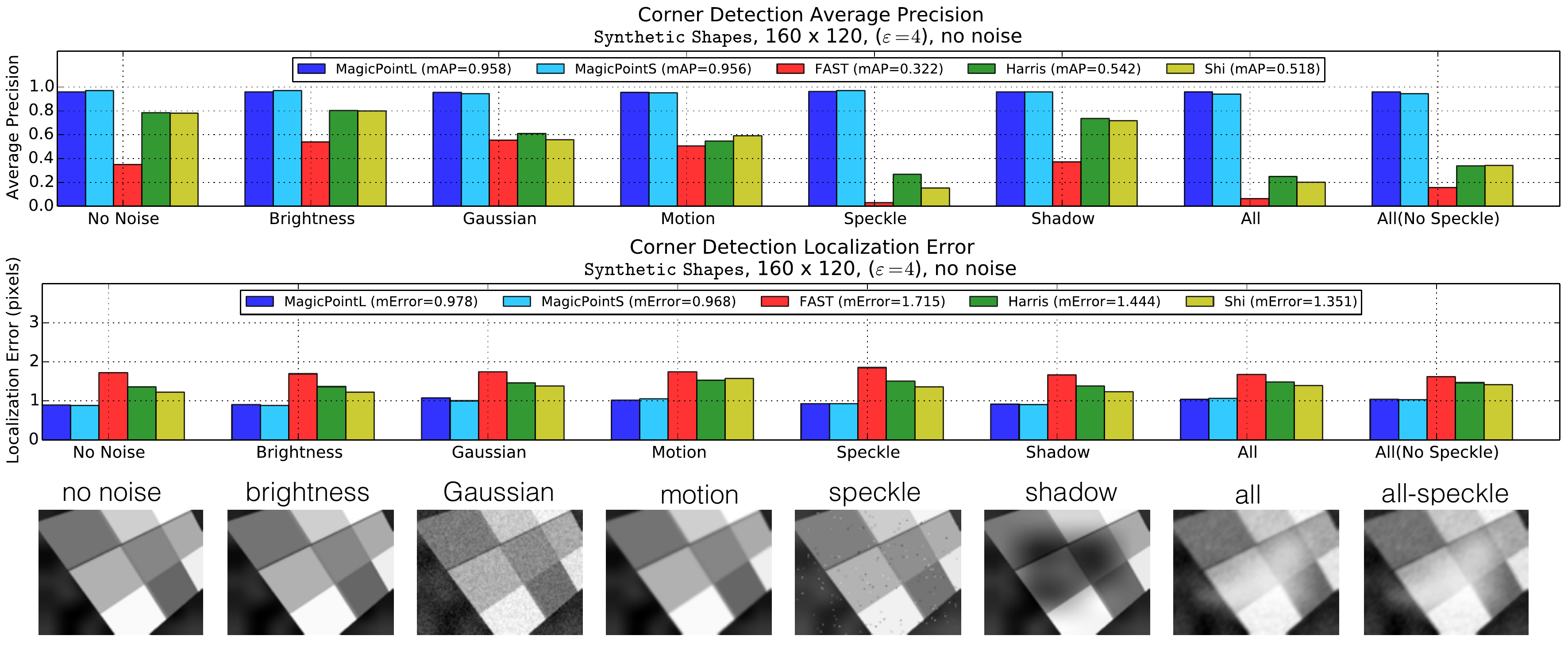}
\caption{{\bf Effect of Noise Type.} The detector performance is broken down by noise category. Speckle noise is particularly difficult for traditional detectors. \label{fig:mpeval_noise_type}} 
\end{figure}

\textbf{Effect of Noise Type.} We categorize the noise into eight categories. We study the effect of these noise types individually to better understand which has the biggest effect on the point detectors. Speckle noise is particularly difficult for traditional detectors. Results are summarized in Figure~\ref{fig:mpeval_noise_type}.

\textbf{Blob Detection.} We experimented with our model's ability to detect the centers of shapes such as quadrilaterals and ellipses. We used the MagicPointL architecture (as described above) and augmented the Synthetic Shapes training set to include blob centers in addition to corners. We observed that our model was able to detect such blobs as long as the entire shape was not too large. However, the confidences produced for such ``blob detection'' are typically lower than those for corners, making it somewhat cumbersome to integrate both kinds of detections into a single system. For the main experiments in the paper, we omit training with blobs, except the following experiment.

\begin{figure}[h]
\centering
\includegraphics[width=.9\linewidth]{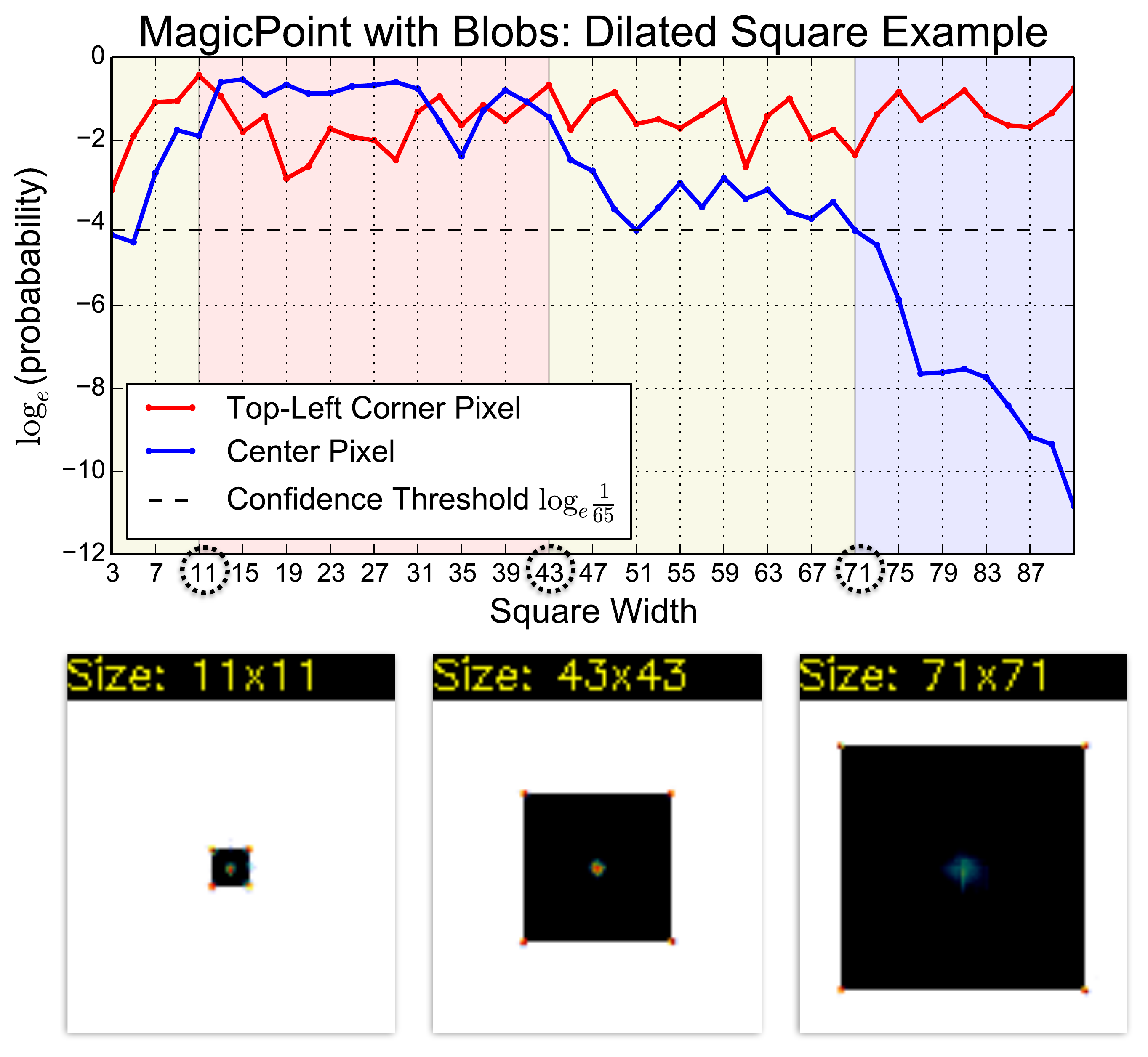}
\caption{{\bf MagicPoint: Blob Center Detection} Top: we experimented with MagicPoint's ability to detect the centers of shapes and plot detection confidences for both the top-left (TL) corner and the center blob. Bottom: point detection heatmaps (MagicPoint outputs) superimposed on the black rectangle images. Notice that our model is able to detect centers of $71$ pixel rectangles, meaning that our network's receptive field is at least $71$ pixels. \label{fig:centers}} 
\vspace{-.15in}
\end{figure}

We created a sequence of $96 \times 96$ images of a black square on a white background. We vary the square's width to range from $3$ to $91$ pixels and report MagicPoint's confidence for two special pixels in the output heatmap: the center pixel (location of the blob) and the square's top-left pixel (an easy-to-detect corner). The MagicPoint blob+corner confidence plot for this experiment can be seen in Figure~\ref{fig:centers}. We observe that we can confidently detect the center of the blob when the square is between $11$ and $43$ pixels wide (red region in Figure~\ref{fig:centers}), detect with lower confidence when the square is between $43$ and $71$ pixels wide (yellow region in Figure~\ref{fig:centers}), and unable to detect the center blob when the square is larger than $71$ (blue regions in Figure~\ref{fig:centers}). 

\section{Homographic Adaptation Experiment}
\label{sec:extra-ha}
When combining interest point response maps, it is important to differentiate between within-scale aggregation and across-scale aggregation. Real-world images typically contain features at different scales, as some points which would be deemed interesting in a high-resolution images, are often not even visible in coarser, lower resolution images. However, within a single-scale, transformations of the image such as rotations and translations should not make interest points appear/disappear. This underlying multi-scale nature of images has different implications for within-scale and across-scale aggregation strategies. Within-scale aggregation should be similar to computing the intersection of a set and across-scale aggregation should be similar to the union of a set. In other words, it is the average response within-scale that we really want, and the maximum response across-scale. We can additionally use the average response across scale as a multi-scale measure of interest point confidence. The average response across scales will be maximized when the interest point is visible across all scales, and these are likely to be the most robust interest points for tracking applications.

\textbf{Within-scale aggregation.} We use the average response across a large number of Homographic warps of the input image. Care should be taken in choosing random homographies because not all homographies are realistic image transformations. The number of homographic warps $N_h$ is a hyper-parameter of our approach. We typically enforce the first homography to be equal to identity, so that $N_h=1$ in our experiments corresponds to doing no homographies (or equivalently, applying the identity Homography). Our experiments range from ``small'' $N_h=10$, to ``medium'' $N_h=100$, and ``large'' $N_h=1000$.

\textbf{Across-scale aggregation.} When aggregating across scales, the number of scales considered $N_s$ is a hyper-parameter of our approach. The setting of $N_s=1$ corresponds to no multi-scale aggregation (or simply aggregating across the large possible image size only). For $N_s>1$, we refer to the multi-scale set of images being processed as ``the multi-scale image pyramid.'' We consider weighting schemes that weigh levels of the pyramid differently, giving higher-resolution images a larger weight. This is important because interest points detected at lower resolutions have poorer localization ability, and we want the final aggregated points to be localized as well as possible.

\begin{figure}
\centering
\includegraphics[width=.95\linewidth]{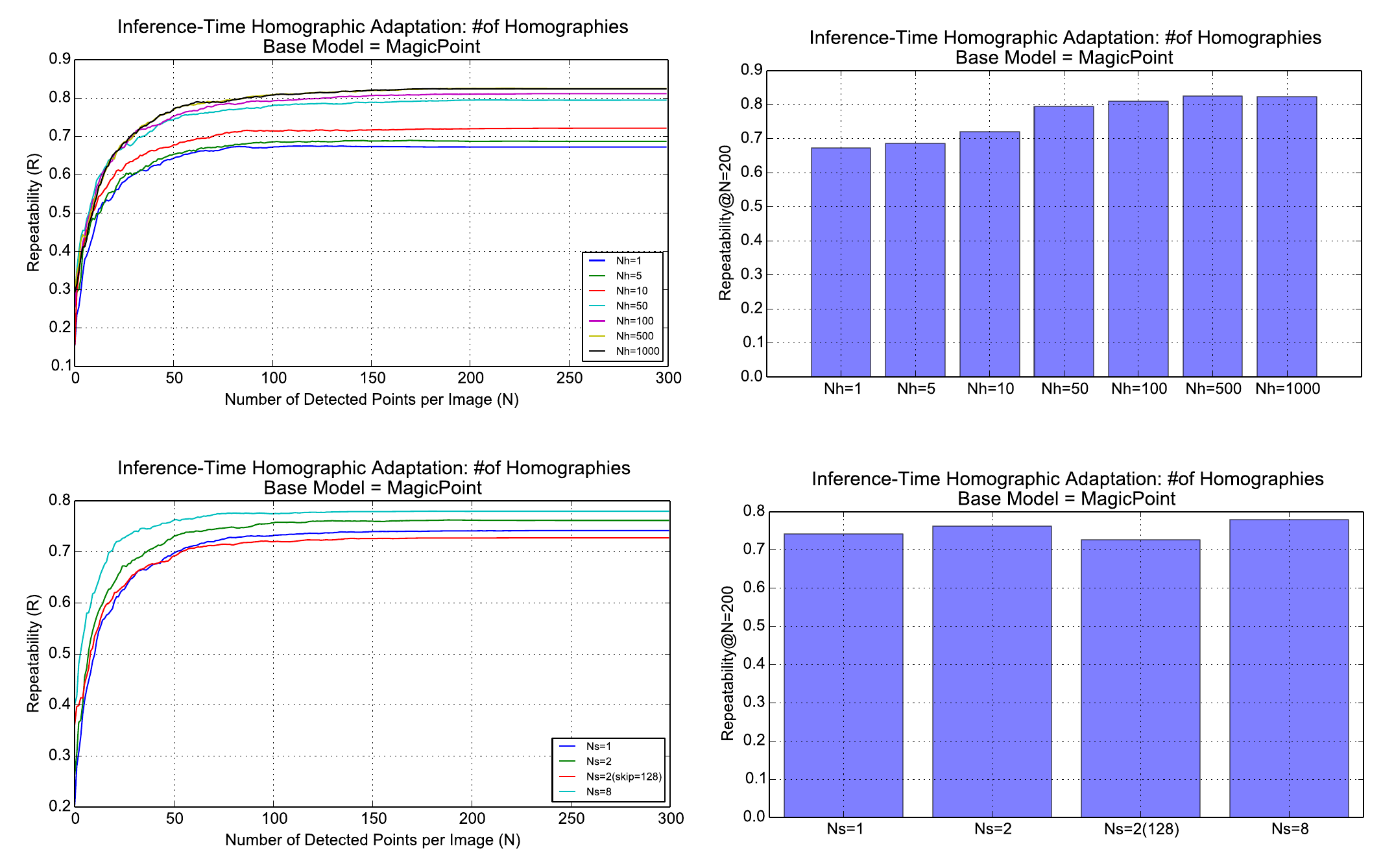}
\caption{\textbf{Homographic Adaptation.} Top: we vary the number of homographies applied during Homographic Adaptation and report repeatability. Bottom: we isolate the effect of scale.} 
\label{fig:ha_h_only_detailed}
\vspace{-.2in}
\end{figure}

We experimented with within-scale and across-scale aggregation on a held out test of MS-COCO images. The results are summarized in Figure~\ref{fig:ha_h_only_detailed}. We find that within-scale aggregation has the biggest effect on repeatability.

\section{Extra Qualitative Examples}
\label{sec:extra-dumps}
\begin{figure*}
\centering
\vspace{-.45in}
\includegraphics[width=\textwidth]{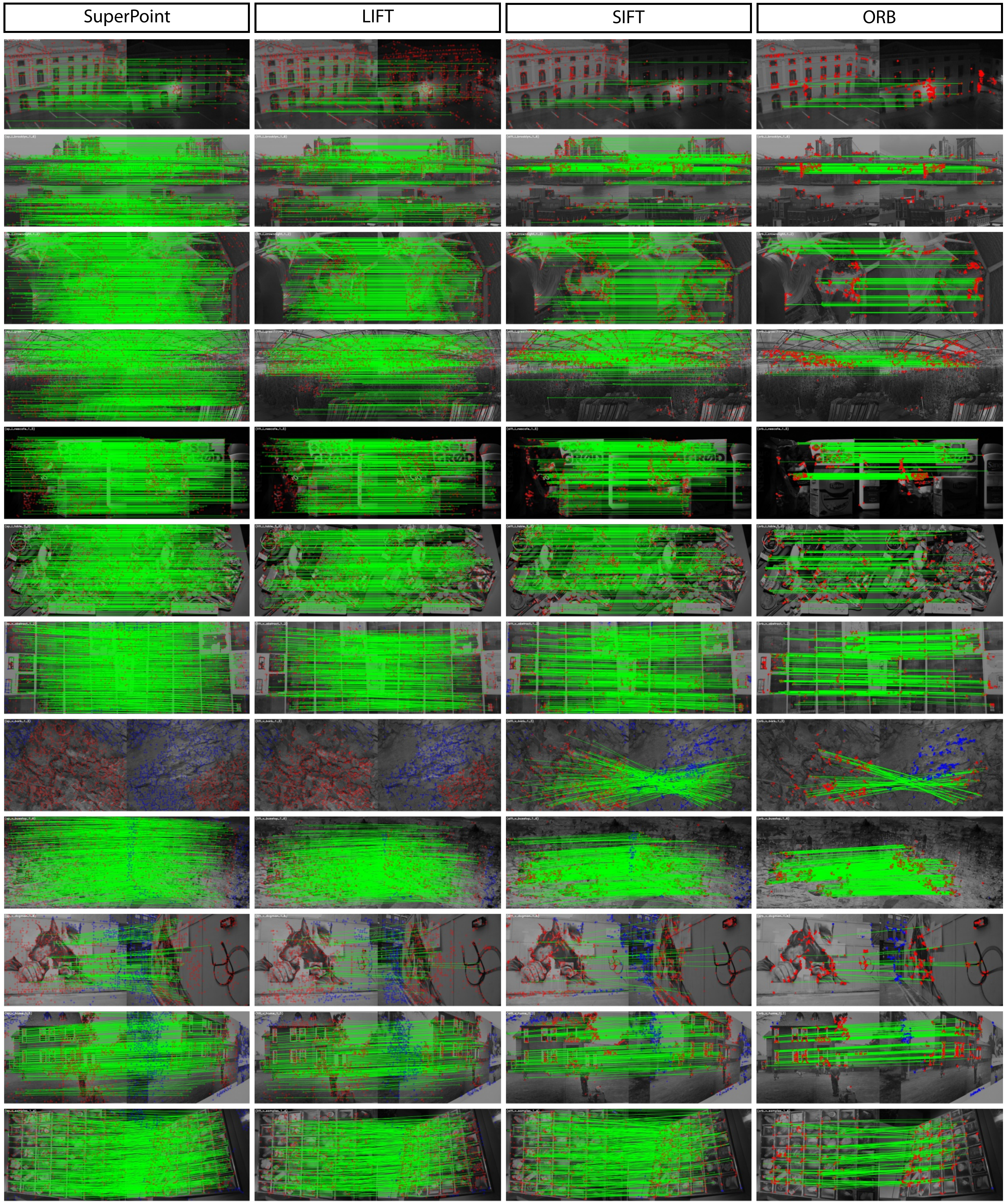}
\caption{{\bf Extra Qualitative Results on HPatches}. More examples like in Figure \ref{h_qualitative}. The green lines show correct correspondences, green dots show matched points, red dots show mis-matched points, blue dots show points outside of the shared viewpoint region.}
\label{fig:extra_dumps}
\end{figure*}

We show extra qualitative examples of SuperPoint, LIFT, SIFT and ORB on HPatches matching in Figure \ref{fig:extra_dumps}.

\end{document}